\documentclass[conference]{IEEEtran}
\usepackage{cite}
\usepackage{amsmath,amssymb,amsfonts}
\usepackage{algorithmic}
\usepackage{graphicx}
\usepackage{textcomp}
\usepackage{xcolor}
\usepackage{multirow}
\newcommand{\sig}{$^{\dagger}$}
\newcommand{\est}[1]{#1}
\def\BibTeX{{\rm B\kern-.05em{\sc i\kern-.025em b}\kern-.08em
    T\kern-.1667em\lower.7ex\hbox{E}\kern-.125emX}}
\begin{document}
\title{CFBrainAug: Modularity-Induced Counterfactual Brain Network Augmentation for Interpretable fMRI-based Autism Diagnosis}

\author{
\IEEEauthorblockN{Yifan Qiao}
\IEEEauthorblockA{
\textit{University of Virginia}\\
Charlottesville, VA, USA\\
tns5yh@virginia.edu}
\and
\IEEEauthorblockN{Sheng Li}
\IEEEauthorblockA{
\textit{University of Virginia}\\
Charlottesville, VA, USA\\
shengli@virginia.edu}
}
\maketitle

\begin{abstract}
Machine learning on resting-state functional magnetic resonance imaging (rs-fMRI) brain networks is promising for diagnosing Autism Spectrum Disorder (ASD), but existing methods are prone to overfitting on limited training data, and their post-hoc statistical attribution offers limited clinical interpretability. We propose CFBrainAug, a counterfactual brain network learning framework that unifies data augmentation and counterfactual explanation by learning label-altered brain networks. Unlike conventional label-preserving augmentation, CFBrainAug first retrieves a label-opposite yet topologically similar candidate and then refines it under a module-number constraint. Jointly training on factual and counterfactual networks enriches the training distribution around the decision boundary, enabling a lightweight feature-selection-based SVM to perform competitively with recent graph deep learning methods. Meanwhile, the counterfactuals provide intrinsic explanations by identifying the connectivity differences between each subject and its counterfactual. Experiments on the ABIDE I dataset show that CFBrainAug reaches an accuracy of 74.1\% on the NYU site, suggesting that counterfactual augmentation can complement more complex models in small-sample brain network analysis.
\end{abstract}


\section{Introduction}
Autism Spectrum Disorder (ASD) is a common neurodevelopmental disorder whose early identification is essential for effective intervention \cite{di2014autism}. Resting-state functional magnetic resonance imaging (rs-fMRI) provides a non-invasive way to model the brain as a functional network, in which abnormal connectivity patterns can serve as objective biomarkers \cite{matthews2004functional,di2014autism}. Graph neural networks (GNNs) \cite{li2021braingnn} and Transformers \cite{kan2022brain} have advanced brain network analysis, but they depend on large datasets, which are scarce for fMRI; with limited samples, these models are prone to overfitting.

\begin{figure}[!t]
\centering
\includegraphics[width=0.9\columnwidth,
trim=4.0cm 3.0cm 6.8cm 1.6cm,
clip]{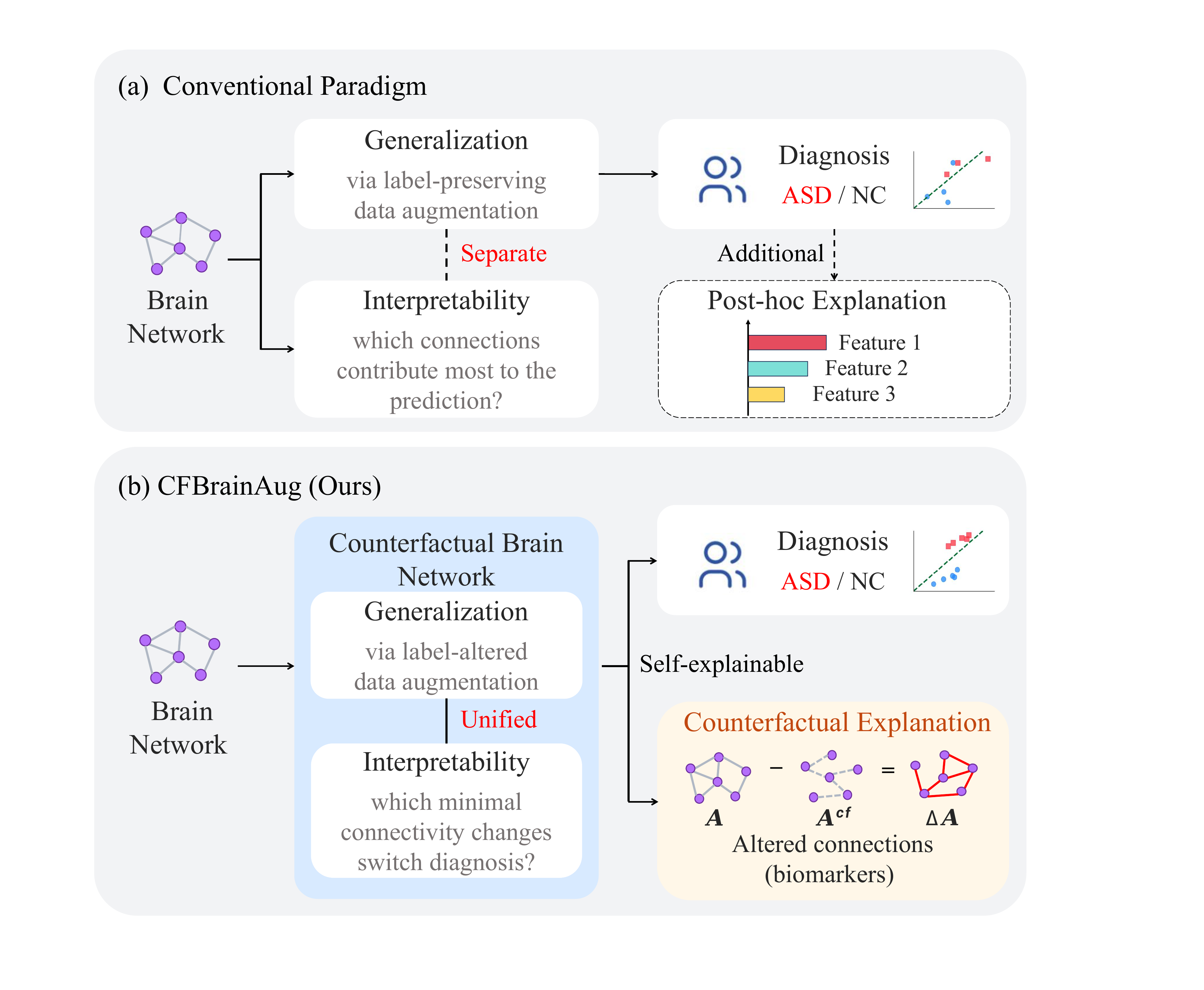}
\caption{Comparison between conventional learning and CFBrainAug. Unlike conventional methods that separately optimize generalization and interpretability, CFBrainAug unifies both through counterfactual brain networks.}
\label{fig:compare}
\end{figure}

Data augmentation is widely used to mitigate this problem \cite{pei2022data}, but existing methods mainly reproduce the observed data distribution \cite{pei2022data,you2020graph,ou2024prior} and may inherit dataset-specific confounders such as age, sex, and scanner \cite{peng2025latent}, which limits their benefit for robust diagnosis \cite{shen2024gcan}. Interpretability is a second obstacle. Current explanation methods identify connections associated with a prediction \cite{mundhenk2019efficient}, are sensitive to model and training settings \cite{kommiya2021towards}, and are applied post hoc, separately from augmentation (Fig.~\ref{fig:compare}). Clinicians, however, are often more interested in which connectivity alterations would change a diagnosis, e.g., from ASD to normal control (NC).

Counterfactual reasoning addresses both issues by asking \textbf{which connectivity alterations would change a diagnostic outcome} \cite{farha2026counterfactual}. A counterfactual brain network is a minimally modified version of the original network that induces a diagnostic label transition while remaining biologically plausible. Such networks enrich the training data with label-transitioned samples and, at the same time, reveal the connectivity changes associated with the transition.

Motivated by this, we propose CFBrainAug, a counterfactual brain network learning framework for generalizable and self-explainable ASD diagnosis. Jointly learning from factual and counterfactual networks enriches the training distribution around the decision boundary, and the factual--counterfactual difference provides an explanation without post-hoc attribution. Our contributions are threefold:
(1) We propose CFBrainAug, which generates label-altered counterfactual brain networks that serve both as augmented training samples and as intrinsic explanations for ASD diagnosis.
(2) We design a counterfactual refinement mechanism with a module-number prior moving counterfactuals towards neurobiological plausibility by preserving the modular organization of brain networks.
(3) Experiments on three ABIDE I sites show that CFBrainAug achieves 74.1\% accuracy on the NYU site, and the connections that differ most between factual and counterfactual networks involve regions repeatedly reported in ASD studies.

\section{Related Work}
\subsection{Data Augmentation for Brain Disease Diagnosis}
Data augmentation is widely used to alleviate overfitting in fMRI-based diagnosis with limited data. Signal-level methods perturb or segment BOLD time series before connectivity estimation \cite{pei2022data,oh2020time}, and graph-level methods drop nodes, perturb edges, or crop subgraphs of brain networks \cite{you2020graph}. Generative models, such as diffusion models, learn the data distribution to synthesize new imaging data \cite{ou2024prior}, and GCAN generates functional connectivity of the target class to guide a diagnostic model \cite{shen2024gcan}. Except for the latter, these methods are label-preserving: they mainly model intra-class variability and provide little information about how a brain network changes across the diagnostic boundary.

\subsection{Interpretability for Brain Disease Diagnosis}
Clinical use of brain disease diagnosis requires biologically meaningful biomarkers in addition to accurate predictions. Post-hoc methods, such as saliency maps \cite{mundhenk2019efficient}, gradient-based attribution, and Shapley-value explanations \cite{kommiya2021towards}, identify important ROIs or connections after training, but they are not part of the model and their results can differ across attribution methods. Intrinsically interpretable models build explanation into learning, for example through ROI-selection pooling in BrainGNN \cite{li2021braingnn} or modularity priors on brain networks \cite{wang2024leveraging,du2023accurate}. Both families mainly indicate which features are important for a prediction, rather than how the connectivity of a subject would have to change for the diagnosis to change.

\subsection{Counterfactual Learning}
Counterfactual learning studies how minimal input modifications alter a prediction \cite{farha2026counterfactual}. Counterfactual generative methods synthesize counterfactual images by recombining separately generated shape, texture, and background \cite{sauer2021counterfactual}, a decomposition with no direct counterpart in brain networks. On graphs, CF-GNNExplainer \cite{lucic2022cf} finds the fewest edge deletions that change a GNN prediction, other methods generate label-altered graphs by edge, node, or structure edits \cite{guo2025counterfactual}, and recent work adds topology-preserving spectral objectives \cite{prenkaj2025graph}. For brain networks, Abrate and Bonchi \cite{abrate2021counterfactual} search for the closest graph assigned to the other class as a post-hoc explanation. These methods use counterfactuals to explain a trained model rather than to train it, and none constrains modular organization; CFBrainAug uses counterfactuals both as training samples and as explanations, and adds a module-number prior.

\section{Methodology}
\subsection{Preliminary Concepts}
\textbf{Definition 1 (Counterfactual Brain Network).}
Given a brain network $G$ with diagnostic label $f(G)$, a counterfactual brain network $G^{cf}$ is obtained by minimally modifying $G$ such that the diagnostic prediction changes:
$G^{cf}=\mathop{\arg\min}\limits_{G'}\;d(G,G')
\ \text{s.t.}\ f(G')\neq f(G)$ and $G'$ satisfies $\mathcal{P}$, where $d(\cdot,\cdot)$ is a distance between brain networks, $f(\cdot)$ is the diagnostic model, and $\mathcal{P}$ denotes the prior constraints for valid brain networks.

\textbf{Definition 2 (Counterfactual Explanation).}
Given $G$ and its counterfactual $G^{cf}$, with adjacency matrices $A$ and $A^{cf}$, the connectivity difference $\Delta A=A^{cf}-A$ constitutes a counterfactual explanation, describing how the network would differ under the other diagnostic label (e.g., ASD$\rightarrow$NC or NC$\rightarrow$ASD).

\begin{figure*}[!t]
\centering
\includegraphics[width=0.8\textwidth, trim=151 67 150 65, clip]{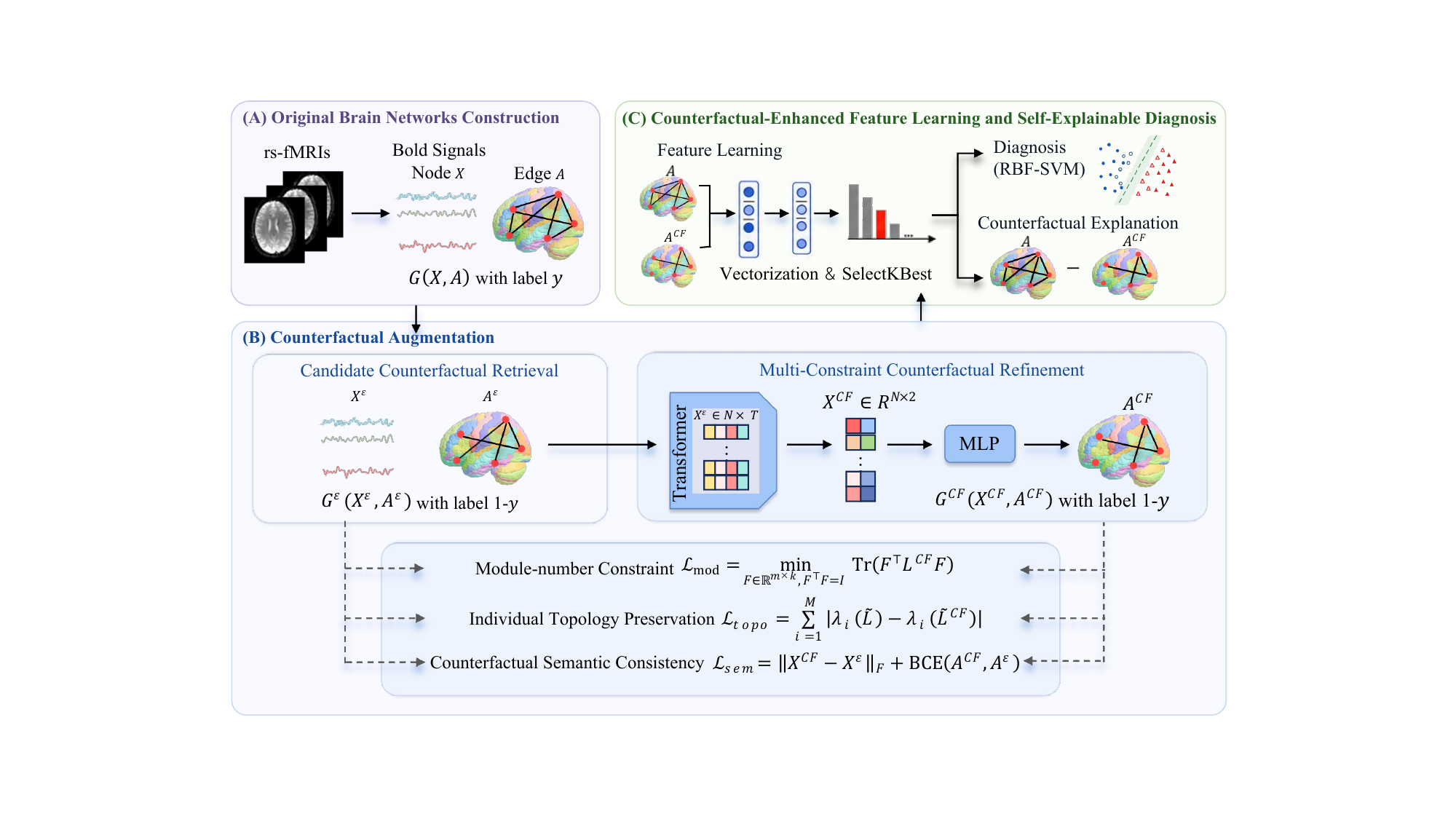}
\caption{Overview of CFBrainAug. (A) Original brain network construction from rs-fMRI. (B) Candidate counterfactual brain networks are retrieved and refined through multi-constraint optimization. (C) Original and counterfactual brain networks are jointly used for RBF-SVM diagnosis, and the most altered connections between factual and counterfactual networks serve as counterfactual explanations.}
\label{fig:overview}
\end{figure*}

\subsection{Original Brain Network Construction}
CFBrainAug consists of three modules (Fig.~\ref{fig:overview}): brain network construction, counterfactual augmentation, and counterfactual-enhanced diagnosis. For each subject, we construct a brain network $G(X, A)$, where $X \in \mathbb{R}^{m \times t}$ contains the BOLD time series of $m$ ROIs with length $t$. The adjacency matrix $A \in \mathbb{R}^{m \times m}$ is defined as the absolute value of the Pearson correlation coefficient, $A_{ij}=|r_{ij}|$, so that all edge weights are non-negative and the resulting graph Laplacians are positive semidefinite. The dataset is denoted as $\{G_i(X_i, A_i)\}_{i=1}^{n}$ with $n$ subjects.

\subsection{Plausible Counterfactual Augmentation}
This module retrieves label-flipped candidates from the training data and refines them with a multi-constraint end-to-end architecture.

\subsubsection{Candidate Counterfactual Retrieval}
Given $G(X,A)$ with label $y$, we retrieve a candidate $G^{\varepsilon}(X^{\varepsilon},A^{\varepsilon})$ among training samples with the opposite label $1-y$, selecting the one closest to $G$ in terms of graph Laplacian similarity \cite{dwivedi2023benchmarking}. Since all experiments are run per site, the candidate always comes from the same site and acquisition protocol as the original subject. $G^{\varepsilon}$ initializes the refinement module.

\subsubsection{Multi-Constraint Counterfactual Refinement}
A refinement module then transforms each retrieved candidate into a subject-specific, plausible counterfactual. Given the strong inter-subject heterogeneity and subtle disease-related connectivity alterations in functional brain networks \cite{li2009review}, the refinement is jointly guided by three complementary constraints: \textbf{module-number constraint, individual topology preservation, and counterfactual semantic consistency}. 

\paragraph{Module-Number Constraint}
Brain networks exhibit dense intra-module and sparse inter-module connectivity \cite{wang2024leveraging}. A graph with non-negative weights has at least $k$ connected components if and only if the $k$ smallest eigenvalues of its Laplacian are zero \cite{chung1997spectral}. We therefore encourage a $k$-module structure by penalizing the sum of the $k$ smallest eigenvalues of the Laplacian $L^{cf}$ of the refined counterfactual, which by Ky Fan's theorem \cite{fan1949weyl} equals a trace minimization:
\begin{equation}
\mathcal{L}_{mod} = \min_{F \in \mathbb{R}^{m \times k},\,F^\top F = I} \mathrm{Tr}\left( F^\top L^{cf} F \right),
\end{equation}
where $F$ contains the eigenvectors corresponding to the smallest $k$ eigenvalues.

\paragraph{Individual Topology Preservation}
To ensure the refined counterfactual remains faithful to the original subject, we constrain it to preserve subject-specific topology through graph Laplacian spectral similarity:
\begin{equation}
\mathcal{L}_{topo} = \sum_{i=1}^{M} \left| \lambda_i(\tilde{L}) - \lambda_i(\tilde{L}^{cf}) \right|,
\end{equation}
where $\tilde{L}$ and $\tilde{L}^{cf}$ are the normalized Laplacians of $G$ and $G^{cf}$, and $\lambda_i(\cdot)$ denotes the $i$-th eigenvalue. Unlike local edge/node perturbations, this constraint focuses on global topological characteristics encoded by eigenvalues, anchoring the counterfactual to the individual's unique topology and preventing drift towards other subjects.

\paragraph{Counterfactual Semantic Consistency}
To retain label-related semantics, the refined counterfactual is kept close to its label-flipped candidate with $1-y$ at the node and edge levels:
\begin{equation}
\mathcal{L}_{sem} = \| X^{cf} - X^{\varepsilon} \|_F + \mathrm{BCE}(A^{cf}, A^{\varepsilon}),
\end{equation}
where $\|\cdot\|_F$ denotes the Frobenius norm for the node feature discrepancy, and $\mathrm{BCE}(\cdot,\cdot)$ denotes the binary cross-entropy to preserve edge distribution consistency. BCE is well defined since both $A^{cf}$ (sigmoid outputs) and $A^{\varepsilon}$ ($A_{ij}=|r_{ij}|$) lie in $[0,1]$.

\paragraph{Overall Refinement Objective}
The three complementary objectives are jointly optimized in an end-to-end Transformer--MLP architecture. Specifically, $G^{\varepsilon}$ is fed to Transformer layers to learn the node feature matrix $X^{cf}$, and an MLP takes paired node features as input to predict edge probabilities, yielding the refined adjacency matrix $A^{cf}$. The overall objective is
\begin{equation}
\mathcal{L}_{refi} = \mathcal{L}_{mod} + \lambda_1 \mathcal{L}_{topo} + \lambda_2 \mathcal{L}_{sem},
\end{equation}
where $\lambda_1$ and $\lambda_2$ balance the three terms. Eq.~(4) contains no classifier term: the label change comes from the opposite-label initialization $G^{\varepsilon}$ and is maintained by $\mathcal{L}_{sem}$, while $\mathcal{L}_{topo}$ keeps the counterfactual close to $G$ (a strategy also explored in \cite{prenkaj2025graph}). Minimizing $\mathcal{L}_{refi}$ thus yields a structurally plausible, individual-preserving, and semantically consistent counterfactual $G^{cf}(X^{cf},A^{cf})$, exposing the model to label-altered connectivity patterns for a more robust decision boundary.

\subsection{Counterfactual-Enhanced Feature Learning and Self-Explainable Diagnosis}
The refined counterfactuals are added to the original networks to form an augmented training set. The two sample types are balanced by the sample weights
\begin{equation}
\mathcal{W}_{total} = \mathcal{W}_{orig} + \lambda \mathcal{W}_{cf},
\end{equation}
where $\mathcal{W}_{orig}$ and $\mathcal{W}_{cf}$ are the weights of factual and counterfactual samples, and $\lambda$ is a balancing hyperparameter. Each adjacency matrix is vectorized by its upper-triangular entries, followed by SelectKBest feature selection and an RBF-SVM. CFBrainAug thus improves generalization by enriching the training data rather than increasing classifier complexity.

CFBrainAug also provides intrinsic counterfactual explanations without an additional post-hoc explanation model. Specifically, counterfactual biomarkers are identified by the magnitude of the connectivity difference in Definition~2,
\begin{equation}
|\Delta A| = |A^{cf}-A|,
\end{equation}
where the most modified functional connections can be regarded as the minimal connectivity alterations needed to reverse the diagnostic prediction, and may serve as candidate biomarkers for further clinical investigation.

\begin{table}[!t]
\caption{Demographics of the ABIDE I subjects. Age: mean (standard deviation); M/F: male/female.}
\label{tab:demo}
\centering
\footnotesize
\setlength{\tabcolsep}{4pt}
\begin{tabular}{lccccc}
\hline
Site & Category & Subject & M & F & Age\\
\hline
\multirow{2}{*}{NYU}
& ASD & 79 & 68 & 11 & 14.5 (6.9) \\
& NC  & 105 & 79 & 26 & 15.8 (6.2) \\
\hline
\multirow{2}{*}{UM}
& ASD & 68 & 58 & 10 & 13.1 (2.4) \\
& NC  & 77 & 59 & 18 & 14.7 (3.5) \\
\hline
\multirow{2}{*}{LEUVEN}
& ASD & 29 & 26 & 3 & 17.7 (5.0) \\
& NC  & 35 & 30 & 5 & 18.1 (4.9) \\
\hline
\end{tabular}
\end{table}

\begin{table*}[!t]
\caption{Performance comparison on the ABIDE I NYU, UM, and LEUVEN sites, reported as mean (standard deviation) over 10 repetitions of 5-fold cross-validation (best in bold; $^{\dagger}$: significantly worse than CFBrainAug, $p<0.05$).}
\label{tab:overall}
\centering
\scriptsize
\setlength{\tabcolsep}{2.5pt}
\renewcommand{\arraystretch}{1.1}
\resizebox{\textwidth}{!}{
\begin{tabular}{l|cccc|cccc|cccc}
\hline
\multirow{2}{*}{Method}
& \multicolumn{4}{c|}{Site NYU}
& \multicolumn{4}{c|}{Site UM}
& \multicolumn{4}{c}{Site LEUVEN} \\
\cline{2-13}
& ACC (\%) & AUC (\%) & SEN (\%) & SPE (\%)
& ACC (\%) & AUC (\%) & SEN (\%) & SPE (\%)
& ACC (\%) & AUC (\%) & SEN (\%) & SPE (\%) \\
\hline
BrainGNN\cite{li2021braingnn}
& 63.9 (5.0)\est{\sig} & 68.1 (5.7)\est{\sig} & 49.0 (8.7)\est{\sig} & 72.0 (9.1)
& 59.4 (12.8)\est{\sig} & 73.6 (9.8) & 58.9 (12.3) & 62.6 (14.9)
& 63.5 (3.7)\est{\sig} & 69.3 (4.5)\est{\sig} & 66.6 (10.6) & 61.7 (9.7)\est{\sig} \\
STGCN\cite{gadgil2020spatio}
& 70.4 (4.8) & 74.3 (4.5) & 67.8 (7.0) & 73.7 (6.3)
& 65.3 (3.9)\est{\sig} & 66.7 (3.6)\est{\sig} & 42.4 (4.2)\est{\sig} & \textbf{84.4 (8.7)}
& 51.4 (2.0)\est{\sig} & 58.3 (7.1)\est{\sig} & 57.9 (13.3) & 45.2 (14.9)\est{\sig} \\
GoFormer\cite{pang2026goformer}
& 72.5 (4.6) & 76.7 (5.1) & 63.9 (6.5) & 79.4 (6.6)
& 62.3 (4.7)\est{\sig} & 72.7 (7.1) & 52.5 (13.9) & 73.8 (7.8)
& 63.7 (5.4)\est{\sig} & 60.4 (6.7)\est{\sig} & 50.6 (7.3) & 74.5 (6.6) \\
Miniformer\cite{li2025miniformer}
& 62.7 (2.6)\est{\sig} & 66.8 (4.6)\est{\sig} & 54.9 (6.4)\est{\sig} & 64.1 (4.1)\est{\sig}
& 67.7 (1.8)\est{\sig} & 69.3 (1.3)\est{\sig} & 60.7 (2.7)\est{\sig} & 74.6 (4.9)
& 66.9 (2.6) & 69.4 (6.2) & 58.9 (8.7) & 78.6 (7.6) \\
\hline
WindowSlicing\cite{oh2020time}
& 69.2 (4.4) & 74.9 (5.3) & 68.8 (9.0) & 70.1 (7.9)
& 72.9 (2.6) & 78.7 (2.7) & \textbf{70.8 (2.3)} & 75.1 (5.6)
& 69.8 (6.2) & 74.6 (7.7) & 63.9 (14.7) & 76.2 (10.9) \\
TimeWarping\cite{oh2020time}
& 65.5 (3.4)\est{\sig} & 72.5 (4.4) & \textbf{71.4 (6.8)} & 60.6 (7.1)\est{\sig}
& 73.3 (1.9) & 78.0 (2.2) & 70.2 (2.6) & 75.8 (5.3)
& 68.3 (7.6) & 75.2 (7.8) & \textbf{71.4 (10.8)} & 67.4 (9.3)\est{\sig} \\
NodeDropping\cite{you2020graph}
& 67.6 (4.6)\est{\sig} & 74.6 (4.9) & 68.9 (6.8) & 67.2 (7.4)\est{\sig}
& 69.3 (3.8)\est{\sig} & 77.4 (2.9) & 65.6 (5.7) & 73.2 (4.6)
& 63.6 (5.6)\est{\sig} & 72.8 (6.3) & 60.1 (13.9) & 67.3 (9.3)\est{\sig} \\
EdgePerturbation\cite{you2020graph}
& 68.8 (4.0)\est{\sig} & 73.3 (5.0) & 66.4 (7.5) & 69.9 (8.0)
& 71.2 (2.0)\est{\sig} & 77.8 (3.0) & 69.6 (4.5) & 72.4 (4.3)\est{\sig}
& 66.7 (6.1) & 73.8 (7.4) & 64.0 (14.7) & 70.5 (11.3) \\
\hline
CF-GNNExplainer\cite{lucic2022cf}
& \est{66.8 (3.9)}\est{\sig} & \est{73.6 (4.8)} & \est{60.2 (7.9)} & \est{72.1 (6.9)}
& \est{69.4 (3.1)}\est{\sig} & \est{74.9 (3.0)} & \est{63.8 (4.6)} & \est{74.2 (5.4)}
& \est{64.2 (5.8)}\est{\sig} & \est{71.5 (7.0)} & \est{59.8 (13.2)} & \est{68.1 (10.4)} \\
Abrate \& Bonchi\cite{abrate2021counterfactual}
& \est{68.1 (3.6)}\est{\sig} & \est{74.5 (4.6)} & \est{61.5 (7.4)} & \est{73.6 (6.5)}
& \est{70.6 (2.8)}\est{\sig} & \est{76.1 (2.9)} & \est{65.1 (4.1)} & \est{75.3 (5.0)}
& \est{65.9 (5.5)} & \est{72.9 (6.8)} & \est{61.6 (12.8)} & \est{69.7 (10.1)} \\
\hline
\textbf{CFBrainAug (Ours)}
& \textbf{74.1 (2.6)} & \textbf{79.2 (4.7)} & 66.7 (6.2) & \textbf{79.5 (6.8)}
& \textbf{75.5 (2.7)} & \textbf{78.9 (2.6)} & 68.5 (2.7) & 81.6 (6.5)
& \textbf{72.4 (4.9)} & \textbf{78.6 (6.5)} & 64.7 (10.8) & \textbf{81.2 (8.7)} \\
\hline
\end{tabular}
}
\end{table*}

\begin{table}[!t]
\centering
\caption{Ablation study on the ABIDE I NYU site (best in bold).}
\label{tab:ablation}
\scriptsize
\setlength{\tabcolsep}{3pt}
\renewcommand{\arraystretch}{1.05}
\resizebox{\columnwidth}{!}{
\begin{tabular}{lcccc}
\hline
Method & ACC (\%) & AUC (\%) & SEN (\%) & SPE (\%) \\
\hline
Full CFBrainAug (Ours) & \textbf{74.1 (2.6)} & \textbf{79.2 (4.7)} & \textbf{66.7 (6.2)} & \textbf{79.5 (6.8)} \\
w/o ModularConstraint & 66.8 (10.6) & 71.0 (14.4) & 65.7 (10.6) & 67.6 (6.2) \\
w/o Refine & 67.4 (9.8) & 71.7 (9.6) & 59.5 (8.7) & 73.3 (12.4) \\
w/o CounterfactualAug & 66.3 (10.7) & 74.0 (10.2) & 65.8 (12.7) & 66.7 (13.6) \\
\hline
\end{tabular}
}
\end{table}

\section{Experiment}
\subsection{Materials and Experimental Settings}
We conduct site-wise ASD diagnosis on three ABIDE I \cite{di2014autism} sites (NYU, UM, and LEUVEN; Table~\ref{tab:demo}). Rs-fMRI data are parcellated into 116 ROIs (AAL) for NYU and LEUVEN and 200 ROIs (CC200) for UM, and the adjacency matrices are constructed from absolute Pearson correlations, $A_{ij}=|r_{ij}|$ with $A_{ii}=0$. We use 10 repetitions of stratified 5-fold cross-validation. The candidate retrieval pool is restricted to the training fold, so no test sample is accessed during retrieval or refinement. We set $k=8$ and $\lambda=1$ in Eq.~(5).

CFBrainAug and all augmentation baselines share the same downstream pipeline: SelectKBest (K=100) followed by an RBF-SVM (C=0.5). We use this lightweight classifier because complex deep models do not consistently improve performance on small functional connectome datasets \cite{han2026rethinking}. Deep learning baselines are trained for 100 epochs with RMSprop (learning rate 0.01), a batch size of 16, and class-balanced binary cross-entropy. We report ACC, AUC, SEN, and SPE. Since all methods share the same 50 test folds, CFBrainAug is compared with each method using the Nadeau--Bengio corrected resampled $t$-test \cite{nadeau2003inference}; all reported $p$-values are Holm--Bonferroni adjusted over the compared methods for each site and metric.

\subsection{Competing Methods}
We compare with four recent brain network models, BrainGNN \cite{li2021braingnn}, STGCN \cite{gadgil2020spatio}, GoFormer \cite{pang2026goformer}, and Miniformer \cite{li2025miniformer}, and four augmentation strategies: Window Slicing and Time Warping \cite{oh2020time}, which segment or temporally distort BOLD signals, and Node Dropping and Edge Perturbation \cite{you2020graph}, which randomly remove 20\% of nodes or edges. We further adapt two counterfactual methods to generate label-altered brain networks for augmentation: CF-GNNExplainer \cite{lucic2022cf}, whose deleted edges are set to zero to flip a GCN trained on each training fold (top 20\% of $A_{ij}$ as edges), and the search of Abrate and Bonchi \cite{abrate2021counterfactual}, applied to weighted networks from the nearest opposite-class training subject. All augmented data are fed into the same feature selection and SVM pipeline as CFBrainAug.

\subsection{Experimental Results}
Table~\ref{tab:overall} contains two kinds of comparison. The augmentation and counterfactual baselines use the same classifier, folds, and hyperparameters as CFBrainAug, so differences in these rows are due to the augmented samples; the four deep models use their own classifiers and serve as reference points. CFBrainAug has the highest ACC and AUC on all three sites, e.g., 74.1\% ACC and 79.2\% AUC on NYU. We attribute this improvement mainly to \textbf{the label-altered counterfactuals, which may better cover the sparsely observed region around the decision boundary than label-preserving random perturbations}. The two counterfactual baselines yield only marginal gains over training without augmentation (66.3\% ACC on NYU), possibly because their minimal edits produce networks that remain close to the original ones despite the altered labels.
CFBrainAug does not have the highest sensitivity on any site, and STGCN has a higher specificity on UM; the bias towards specificity is likely due to the NC-majority class imbalance (Table~\ref{tab:demo}), and improving sensitivity remains future work.

\begin{figure}[!t]
\centering
\includegraphics[width=0.65\columnwidth]{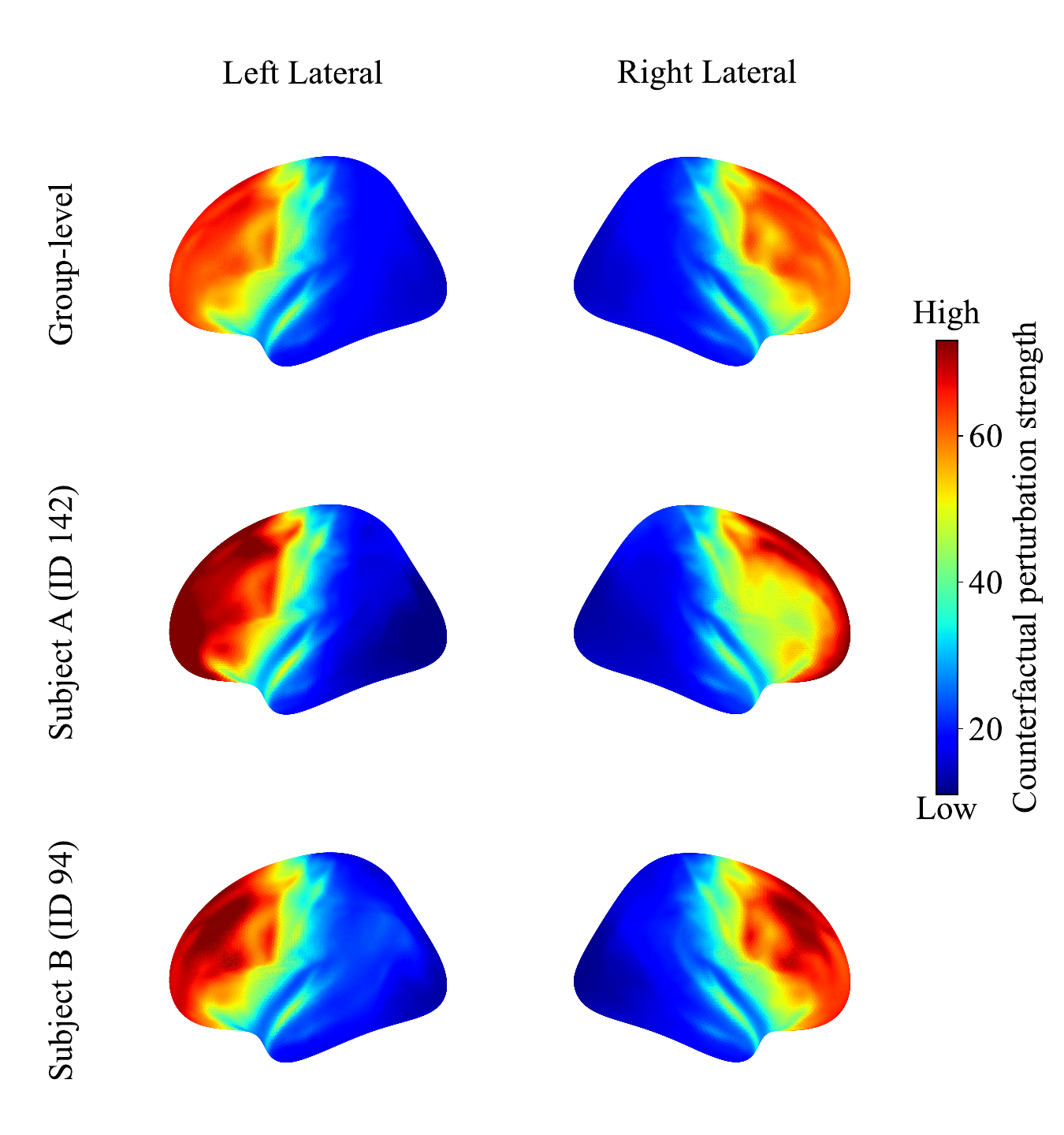}
\caption{Group-level (top) and subject-specific (middle and bottom) counterfactual explanations on the NYU site. The group-level map averages $|\Delta A|$ over all test subjects; the subject-specific maps correspond to two randomly selected correctly classified subjects. Warmer colors indicate larger counterfactual perturbation strength $s_i$ (Sec.~\ref{sec:biomarker}).}
\label{fig:surface}
\end{figure}

\begin{figure}[!t]
\centering
\includegraphics[width=0.72\columnwidth]{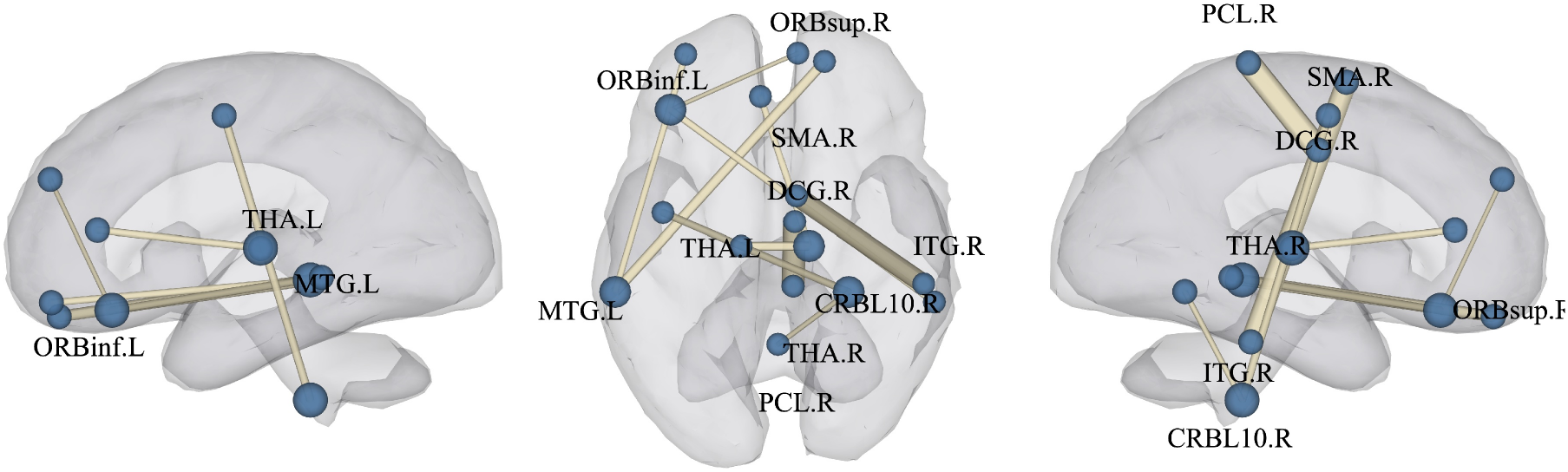}
\caption{Top-10 connections from the group-level counterfactual explanation on the NYU site. Thicker edges indicate larger differences.}
\label{fig:top10}
\end{figure}

\begin{figure}[!t]
\centering
\includegraphics[width=0.9\columnwidth]{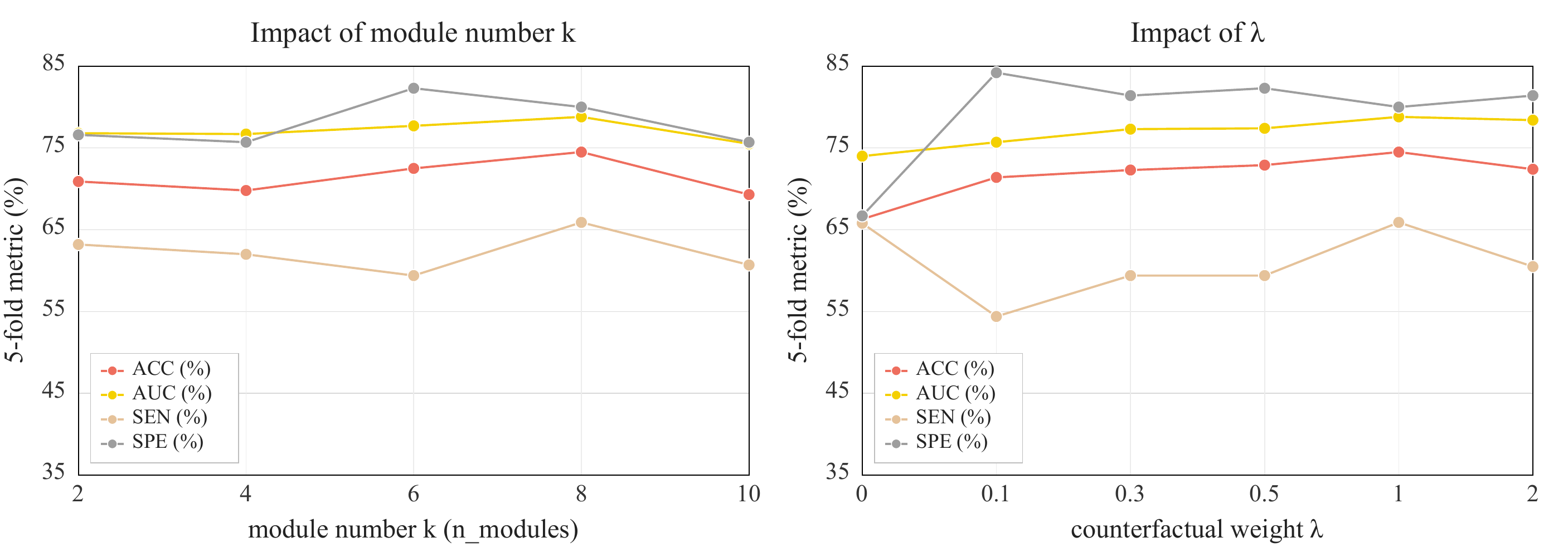}
\caption{Sensitivity to the number of modules $k$ (left) and the counterfactual weight $\lambda$ (right) on the NYU site.}
\label{fig:hyper}
\end{figure}

\subsection{Ablation Study}
We compare the full model with three variants (Table~\ref{tab:ablation}): (1) w/o ModularConstraint, which discards the module-number constraint; (2) w/o Refine, which uses the raw candidates for augmentation; and (3) w/o CounterfactualAug, which trains only on the original data. Removing any component lowers all four metrics, indicating that the modular constraint, the refinement, and the counterfactual augmentation each contribute to CFBrainAug.

\subsection{Counterfactual Quality Assessment}
We use three metrics. Label Flipping Rate (LFR) is the percentage of counterfactuals that the SVM trained on the same training fold assigns to the target class $1-y$; Topological Similarity (TS) is the Laplacian spectral similarity between factual and counterfactual networks; and Module-Number Preservation Rate (MPR) is the percentage of counterfactuals with the prescribed module number. 
CFBrainAug achieves an LFR of 66.9\%, since the classifier-free objective encourages but does not enforce a label flip; a classifier-guided flip term may further improve it. The TS of 87.1\% indicates that the counterfactuals largely preserve the subject-specific global topology. The MPR of 60.0\% suggests that the soft module-number penalty, which trades off against the other two terms in Eq.~(4), is not always satisfied; adaptive loss weighting may further improve it.
\subsection{Counterfactual Biomarker Identification}
\label{sec:biomarker}
Following Eq.~(6), we compute $|\Delta A|$ at the group and subject levels, where the group-level result averages $|\Delta A|$ over all test subjects on the NYU site, and the perturbation strength of ROI $i$ is $s_i=\sum_j |\Delta A_{ij}|$. As shown in Fig.~\ref{fig:surface}, regions with larger $s_i$ mainly include the insula, supplementary motor area (SMA), and precentral gyrus, consistent with previous ASD studies \cite{just2004cortical}. The two subjects additionally show distinct patterns (orbitofrontal cortex for Subject A; middle and superior frontal gyri for Subject B), suggesting that CFBrainAug also captures individual characteristics. The top-10 connections (Fig.~\ref{fig:top10}) involve the sensorimotor network, the default mode network, and subcortical--cerebellar circuits, which have been widely reported in ASD \cite{zhang2026executive,wang2026static}.

\subsection{Hyperparameter Sensitivity Analysis}
As shown in Fig.~\ref{fig:hyper}, ACC and AUC increase with $k$, peak at $k=8$, and then slightly decline, which is consistent with \cite{du2023accurate}, where $k=8$ also gave the best AUC on AAL-116 networks. A small $k$ provides insufficient structural constraints, whereas a large $k$ over-fragments the modules. Increasing $\lambda$ improves performance over $\lambda=0$, peaking at $\lambda=1$.

\section{Conclusion and Future Work}
We propose CFBrainAug, a module-number-guided counterfactual brain network augmentation framework that integrates data augmentation with intrinsic counterfactual explanation for rs-fMRI-based ASD diagnosis. Experiments on three ABIDE I sites show that a simple SVM pipeline with counterfactual augmentation is competitive with recent graph deep learning models, suggesting that counterfactual augmentation can complement more complex models in small-sample brain network analysis. This study is limited to within-site evaluation on three ABIDE I sites (64 to 184 subjects each), and the label flip of the counterfactuals is encouraged rather than guaranteed. Future work will validate CFBrainAug on larger multi-site cohorts, explore classifier-guided refinement, and examine the biological validity of the identified connectivity differences.

\section*{Acknowledgments}
This work is supported in part by the U.S.~Office of Naval Research Award under Grant Number N00014-24-1-2668, the National Science Foundation under Grants IIS-2316306 and CNS-2330215, the National Institutes of Health (NIH) under Grant R01EB293388, Merck BARDS Academic Collaboration Grant, and a 4-VA grant.

\bibliographystyle{IEEEtran}
\bibliography{reference}

\begin{thebibliography}{10}
\providecommand{\url}[1]{#1}
\csname url@samestyle\endcsname
\providecommand{\newblock}{\relax}
\providecommand{\bibinfo}[2]{#2}
\providecommand{\BIBentrySTDinterwordspacing}{\spaceskip=0pt\relax}
\providecommand{\BIBentryALTinterwordstretchfactor}{4}
\providecommand{\BIBentryALTinterwordspacing}{\spaceskip=\fontdimen2\font plus
\BIBentryALTinterwordstretchfactor\fontdimen3\font minus
  \fontdimen4\font\relax}
\providecommand{\BIBforeignlanguage}[2]{{%
\expandafter\ifx\csname l@#1\endcsname\relax
\typeout{** WARNING: IEEEtran.bst: No hyphenation pattern has been}%
\typeout{** loaded for the language `#1'. Using the pattern for}%
\typeout{** the default language instead.}%
\else
\language=\csname l@#1\endcsname
\fi
#2}}
\providecommand{\BIBdecl}{\relax}
\BIBdecl

\bibitem{di2014autism}
A.~Di~Martino, C.-G. Yan, Q.~Li, E.~Denio, F.~X. Castellanos, K.~Alaerts, J.~S.
  Anderson, M.~Assaf, S.~Y. Bookheimer, M.~Dapretto \emph{et~al.}, ``The autism
  brain imaging data exchange: towards a large-scale evaluation of the
  intrinsic brain architecture in autism,'' \emph{Molecular psychiatry},
  vol.~19, no.~6, pp. 659--667, 2014.

\bibitem{matthews2004functional}
P.~M. Matthews and P.~Jezzard, ``Functional magnetic resonance imaging,''
  \emph{Journal of Neurology, Neurosurgery \& Psychiatry}, vol.~75, no.~1, pp.
  6--12, 2004.

\bibitem{li2021braingnn}
X.~Li, Y.~Zhou, N.~Dvornek, M.~Zhang, S.~Gao, J.~Zhuang, D.~Scheinost, L.~H.
  Staib, P.~Ventola, and J.~S. Duncan, ``Braingnn: Interpretable brain graph
  neural network for fmri analysis,'' \emph{Medical image analysis}, vol.~74,
  p. 102233, 2021.

\bibitem{kan2022brain}
X.~Kan, W.~Dai, H.~Cui, Z.~Zhang, Y.~Guo, and C.~Yang, ``Brain network
  transformer,'' \emph{Advances in Neural Information Processing Systems},
  vol.~35, pp. 25\,586--25\,599, 2022.

\bibitem{pei2022data}
S.~Pei, C.~Wang, S.~Cao, and Z.~Lv, ``Data augmentation for fmri-based
  functional connectivity and its application to cross-site adhd
  classification,'' \emph{IEEE Transactions on Instrumentation and
  Measurement}, vol.~72, pp. 1--15, 2022.

\bibitem{you2020graph}
Y.~You, T.~Chen, Y.~Sui, T.~Chen, Z.~Wang, and Y.~Shen, ``Graph contrastive
  learning with augmentations,'' \emph{Advances in neural information
  processing systems}, vol.~33, pp. 5812--5823, 2020.

\bibitem{ou2024prior}
Z.~Ou, C.~Jiang, Y.~Pan, Y.~Zhang, Z.~Cui, and D.~Shen, ``A
  prior-information-guided residual diffusion model for multi-modal pet
  synthesis from mri,'' in \emph{Proceedings of the Thirty-Third International
  Joint Conference on Artificial Intelligence}, 2024, pp. 4769--4777.

\bibitem{peng2025latent}
W.~Peng, T.~Xia, F.~D.~S. Ribeiro, T.~Bosschieter, E.~Adeli, Q.~Zhao,
  B.~Glocker, and K.~M. Pohl, ``Latent causal modeling for 3d brain mri
  counterfactuals,'' in \emph{MICCAI Workshop on Deep Generative Models}.\hskip
  1em plus 0.5em minus 0.4em\relax Springer, 2025, pp. 192--201.

\bibitem{shen2024gcan}
X.~Shen, Z.~Song, and Z.~Zhang, ``Gcan: Generative counterfactual
  attention-guided network for explainable cognitive decline diagnostics based
  on fmri functional connectivity,'' in \emph{International Conference on
  Medical Image Computing and Computer-Assisted Intervention}.\hskip 1em plus
  0.5em minus 0.4em\relax Springer, 2024, pp. 416--426.

\bibitem{mundhenk2019efficient}
T.~N. Mundhenk, B.~Y. Chen, and G.~Friedland, ``Efficient saliency maps for
  explainable ai,'' \emph{arXiv preprint arXiv:1911.11293}, 2019.

\bibitem{kommiya2021towards}
R.~Kommiya~Mothilal, D.~Mahajan, C.~Tan, and A.~Sharma, ``Towards unifying
  feature attribution and counterfactual explanations: Different means to the
  same end,'' in \emph{Proceedings of the 2021 AAAI/ACM Conference on AI,
  Ethics, and Society}, 2021, pp. 652--663.

\bibitem{farha2026counterfactual}
R.~Farha, B.~Ojeme, F.~Khalifa, and M.~M. Rahman, ``Counterfactual,
  longitudinal, and multimodal explainable ai for mri-based alzheimer’s
  diagnosis: A structured review,'' \emph{Journal of Dementia and Alzheimer’s
  Disease}, vol.~3, no.~2, p.~26, 2026.

\bibitem{oh2020time}
C.~Oh, S.~Han, and J.~Jeong, ``Time-series data augmentation based on
  interpolation,'' \emph{Procedia Computer Science}, vol. 175, pp. 64--71,
  2020.

\bibitem{wang2024leveraging}
Q.~Wang, W.~Wang, Y.~Fang, P.-T. Yap, H.~Zhu, H.-J. Li, L.~Qiao, and M.~Liu,
  ``Leveraging brain modularity prior for interpretable representation learning
  of fmri,'' \emph{IEEE Transactions on Biomedical Engineering}, vol.~71,
  no.~8, pp. 2391--2401, 2024.

\bibitem{du2023accurate}
Y.~Du, G.~Wang, C.~Wang, Y.~Zhang, X.~Xi, L.~Zhang, and M.~Liu, ``Accurate
  module induced brain network construction for mild cognitive impairment
  identification with functional {MRI},'' \emph{Frontiers in Aging
  Neuroscience}, vol.~15, p. 1101879, 2023.

\bibitem{sauer2021counterfactual}
A.~Sauer and A.~Geiger, ``Counterfactual generative networks,'' in
  \emph{International Conference on Learning Representations (ICLR)}, 2021.

\bibitem{lucic2022cf}
A.~Lucic, M.~A. ter Hoeve, G.~Tolomei, M.~de~Rijke, and F.~Silvestri,
  ``{CF-GNNExplainer}: Counterfactual explanations for graph neural networks,''
  in \emph{Proceedings of the 25th International Conference on Artificial
  Intelligence and Statistics (AISTATS)}, ser. PMLR, vol. 151, 2022, pp.
  4499--4511.

\bibitem{guo2025counterfactual}
Z.~Guo, Z.~Wu, T.~Xiao, C.~Aggarwal, H.~Liu, and S.~Wang, ``Counterfactual
  learning on graphs: A survey,'' \emph{Machine Intelligence Research},
  vol.~22, no.~1, pp. 17--59, 2025.

\bibitem{prenkaj2025graph}
B.~Prenkaj, E.~Zaradoukas, and G.~Kasneci, ``Graph inverse style transfer for
  counterfactual explainability,'' \emph{arXiv preprint arXiv:2505.17542},
  2025.

\bibitem{abrate2021counterfactual}
C.~Abrate and F.~Bonchi, ``Counterfactual graphs for explainable classification
  of brain networks,'' in \emph{Proceedings of the 27th ACM SIGKDD Conference
  on Knowledge Discovery and Data Mining}, 2021, pp. 2495--2504.

\bibitem{dwivedi2023benchmarking}
V.~P. Dwivedi, C.~K. Joshi, A.~T. Luu, T.~Laurent, Y.~Bengio, and X.~Bresson,
  ``Benchmarking graph neural networks,'' \emph{Journal of Machine Learning
  Research}, vol.~24, no.~43, pp. 1--48, 2023.

\bibitem{li2009review}
K.~Li, L.~Guo, J.~Nie, G.~Li, and T.~Liu, ``Review of methods for functional
  brain connectivity detection using fmri,'' \emph{Computerized medical imaging
  and graphics}, vol.~33, no.~2, pp. 131--139, 2009.

\bibitem{chung1997spectral}
F.~R. Chung, \emph{Spectral graph theory}.\hskip 1em plus 0.5em minus
  0.4em\relax American Mathematical Soc., 1997, vol.~92.

\bibitem{fan1949weyl}
K.~Fan, ``On a theorem of weyl concerning eigenvalues of linear transformations
  i,'' \emph{Proceedings of the National Academy of Sciences}, vol.~35, no.~11,
  pp. 652--655, 1949.

\bibitem{gadgil2020spatio}
S.~Gadgil, Q.~Zhao, A.~Pfefferbaum, E.~V. Sullivan, E.~Adeli, and K.~M. Pohl,
  ``Spatio-temporal graph convolution for resting-state fmri analysis,'' in
  \emph{International conference on medical image computing and
  computer-assisted intervention}.\hskip 1em plus 0.5em minus 0.4em\relax
  Springer, 2020, pp. 528--538.

\bibitem{pang2026goformer}
M.~Pang, L.~Zhou, X.~Yao, J.~Yang, J.~Zhang, Y.~Zhang, L.~Zhang, and L.~Qiao,
  ``Goformer: A golpp inspired transformer for functional brain graph learning
  and classification,'' \emph{Neural networks: the official journal of the
  International Neural Network Society}, vol. 193, p. 108081, 2026.

\bibitem{li2025miniformer}
Y.~Li, J.~Yang, M.~Pang, S.~Zhang, C.~Nugent, M.~Liu, and L.~Qiao,
  ``Miniformer: A minimalist transformer for brain functional networks
  analysis,'' \emph{IEEE Journal of Biomedical and Health Informatics}, 2025.

\bibitem{han2026rethinking}
K.~Han, Y.~Su, L.~He, L.~Zhan, S.~Plis, V.~Calhoun, and C.~Yang, ``Rethinking
  functional brain connectome analysis: do graph deep learning models help,''
  \emph{npj Artificial Intelligence}, vol.~2, no.~1, p.~19, 2026.

\bibitem{nadeau2003inference}
C.~Nadeau and Y.~Bengio, ``Inference for the generalization error,''
  \emph{Machine Learning}, vol.~52, no.~3, pp. 239--281, 2003.

\bibitem{just2004cortical}
M.~A. Just, V.~L. Cherkassky, T.~A. Keller, and N.~J. Minshew, ``Cortical
  activation and synchronization during sentence comprehension in
  high-functioning autism: evidence of underconnectivity,'' \emph{Brain}, vol.
  127, no.~8, pp. 1811--1821, 2004.

\bibitem{zhang2026executive}
X.~Zhang, G.~Qiu, Z.~Mou, C.~Chen, J.~Lu, Z.~Tang, and Z.~Zhou, ``Executive
  dysfunction in autism spectrum disorder is associated with increased
  cerebro-cerebellar resting-state functional connectivity,'' \emph{Neural
  Plasticity}, vol. 2026, no.~1, p. 7449692, 2026.

\bibitem{wang2026static}
M.~Wang, L.~Ma, X.~Li, J.~Huang, and Z.~Wan, ``Static-dynamic functional
  connectivity fusion via cross-attention for asd identification,''
  \emph{Biomedical Signal Processing and Control}, vol. 112, p. 108427, 2026.

\end{thebibliography}

\end{document}